\documentclass[10pt,twocolumn,letterpaper]{article}

\usepackage{wacv}              

\usepackage{graphicx}
\usepackage{amsmath}
\usepackage{amssymb}
\usepackage{booktabs}

\usepackage[pagebackref,breaklinks,colorlinks]{hyperref}

\usepackage[capitalize]{cleveref}
\crefname{section}{Sec.}{Secs.}
\Crefname{section}{Section}{Sections}
\Crefname{table}{Table}{Tables}
\crefname{table}{Tab.}{Tabs.}

\def\confName{WACV}
\def\confYear{2025}

\begin{document}

\title{Automatic Temporal Segmentation for Post-Stroke Rehabilitation: A Keypoint Detection and Temporal Segmentation Approach for Small Datasets }


\author{Jisoo Lee\\
Arizona State University\\
{\tt\small jlee815@asu.edu}
\and
Tamim Ahmed\\
University of Southern California\\
{\tt\small tamimahm@usc.edu}
\and
Thanassis Rikakis\\
University of Southern California\\
{\tt\small rikakis@usc.edu}
\and
Pavan Turaga\\
Arizona State University\\
{\tt\small Pavan.Turaga@asu.edu}
}
\maketitle

\begin{abstract}
   Rehabilitation is essential and critical for post-stroke patients, addressing both physical and cognitive aspects. Stroke predominantly affects older adults, with 75\% of cases occurring in individuals aged 65 and older, underscoring the urgent need for tailored rehabilitation strategies in aging populations. Despite the critical role therapists play in evaluating rehabilitation progress and ensuring the effectiveness of treatment, current assessment methods can often be subjective, inconsistent, and time-consuming, leading to delays in adjusting therapy protocols. 
   This study aims to address these challenges by providing a solution for consistent and timely analysis. Specifically, we perform temporal segmentation of video recordings to capture detailed activities during stroke patients' rehabilitation. The main application scenario motivating this study is the clinical assessment of daily tabletop object interactions, which are crucial for post-stroke physical rehabilitation. 
   To achieve this, we present a framework that leverages the biomechanics of movement during therapy sessions. Our solution divides the process into two main tasks: 2D keypoint detection to track patients' physical movements, and 1D time-series temporal segmentation to analyze these movements over time. This dual approach enables automated labeling with only a limited set of real-world data, addressing the challenges of variability in patient movements and limited dataset availability. By tackling these issues, our method shows strong potential for practical deployment in physical therapy settings, enhancing the speed and accuracy of rehabilitation assessments.
\end{abstract}

\vfill

\section{Introduction}
\label{sec:intro}

Stroke is a medical condition that occurs when the blood supply to the brain is interrupted, resulting in brain tissue damage, which can lead to disability or even death. Aging is a major contributor to stroke, with the risk doubling every decade after the age of 55, and approximately 75\% of stroke patients being 65 or older\cite{Yousufuddin2019}. Rehabilitation therapy is essential for minimizing disability and helping patients recover physical and cognitive functions. This need is especially critical for older adults, as aging leads to greater vulnerability in both physical and cognitive functions, making rehabilitation even more necessary.

Effective rehabilitation relies on thorough assessments to tailor treatment plans. Currently, therapists monitor these assessments, evaluating the patient's physical and cognitive progress to refine therapy protocols accordingly. However, this process is time-intensive and may vary among therapists, creating a need for more objective, automated solutions.

Advancements in deep learning have enabled automation across various fields, including healthcare. In rehabilitation, deep learning offers the potential to streamline and enhance the assessment process. Despite this potential, challenges remain, including the limited availability of real-world patient data, difficulties in using synthetic data, and the computational demands of processing video data, which often involve spatial and temporal complexities.

To overcome these challenges, we propose a novel framework that decomposes complex tasks into smaller, more manageable sub-tasks based on domain-specific insights. By focusing on critical hand-object interactions during the Action Research Arm Test (ARAT), we isolate key movements and extract 2D joint coordinates for detailed analysis. This targeted approach allows us to reduce model complexity and mitigate overfitting, even with a small dataset. We utilize the ASAR (Affective State for ARAT Rehab) dataset \cite{AhmedICMI2023}, which consists of video recordings of stroke patients performing the standardized Action Research Arm Test (ARAT) \cite{YozbatiranNeuro2008}. ARAT assesses upper extremity motor function in stroke patients by evaluating their ability to perform tasks such as grasping, moving, and lifting objects, aiding in tracking rehabilitation progress.

Additionally, given the subjective nature of the ARAT assessment, the criteria for segmenting actions may vary depending on the therapist. To address this, instead of retraining the entire model, we propose adjusting only the temporal segmentation phase to accommodate the changed criteria. This approach offers significant flexibility by allowing adaptation to different segmentation standards without the need for complete model retraining.

\subsection{Small Data Statement}
This study leverages video recordings of 106 stroke patients performing 19 different tasks as part of the Action Research Arm Test (ARAT) assessment. The study involved training a segmentation model to identify specific timestamps in videos captured from three different perspectives: top, left, and right views. After excluding 7 tasks without video recordings and considering cases with excessive occlusion that could not be used for training, the final dataset consisted of 561, 643, and 319 data points for the top, ipsilateral, and contralateral views, respectively. Due to the real-world nature of the data, involving actual stroke patients, augmenting the dataset with time-based scaling or other data augmentation techniques could significantly affect critical aspects such as task completion time, which are important for accurate assessment. Therefore, no augmentation techniques were applied. Additionally, given the focus on capturing subtle hand movements of stroke patients, the use of synthetic data was also deemed inappropriate for this study. Due to the limitations in data size and the need for realistic, patient-specific details, this study is classified as small data research.

In this study, the primary challenge is the small amount of data, in addition to the fact that the data consists of videos, which have high dimensionality. Moreover, due to the nature of the ARAT, there is a need to capture the relationship between the patient's fingers and the target object’s movement within a very narrow range, which presents further difficulties. Due to the small dataset size, even when utilizing pre-trained models, there are significant challenges, as the characteristics of the video data differ from those of existing datasets. Furthermore, applying complex models with a large number of parameters, such as a 3D CNN\cite{Tran2015}
, to such a limited dataset can lead to issues like overfitting.

To address dataset limitations and resource constraints, we propose a new framework that breaks down a large task into smaller, more manageable sub-tasks, based on a detailed understanding of the collected data. Our prior knowledge includes the fact that the patient always interacts with a single object and is consistently instructed to move the object from the table to a shelf above. Therefore, it is more efficient to focus solely on this movement, rather than the entire video frame, to estimate the timestamp of the patient's action label. By concentrating on the target objects and the patients' hand movements, we extract the 2D coordinates of key body joints from each video. We refine these results and apply a transformer-based model \cite{VaswaniNeurIPS2017} for time-series segmentation to predict the action segments. This approach mitigates the risk of overfitting associated with small datasets and demonstrates effective action segmentation using limited computational resources, highlighting its potential for real-world deployment.

\section{Related Work}
Although various studies have introduced methods for video action segmentation \cite{JiPAMI2010, Wang2016ECCV, Neimark2021, Jiang2023}, these methods typically rely on direct use of video data, often from extensive benchmark datasets. Additionally, since these methods are designed to infer the final labels directly from the video, they face challenges in providing intermediate process information to therapists. Furthermore, if the criteria for segmentation change, the entire dataset must be retrained, posing a significant drawback.

Moreover, the datasets used in these studies typically involve labeling based on the overall assessment of human actions, rather than distinguishing fine hand movements. This makes it challenging to apply these methods to our real-world data. Furthermore, in the ASAR dataset, learning relationships between entire frames is inefficient since areas of interest are predefined, whereas models trained on benchmark datasets typically consider the entire frame, including backgrounds. To address this, we divide the task into 2D object detection and temporal segmentation. For hand landmark detection, we use Google's MediaPipe \cite{Lugaresi2019}, and for object detection, among various CNN-based models \cite{LinCVPR2017, LiICCV2019} and transformer-based models \cite{CarionECCV2020}, we choose TridentNet \cite{LiICCV2019} for its computational efficiency achieved through weight-sharing mechanisms.

For temporal segmentation, statistical methods like cusum \cite{Page1954} and Bayesian change point detection \cite{adams2007bayesian} are available but they are complex and sensitive to noise. In contrast, deep-learning models 
 \cite{Hochreiter1997, LeaCVPR2017, Cho-etal-2014-learning, VaswaniNeurIPS2017} offer superior flexibility, robustness, and scalability. We focus on transformers for their ability to capture the overall context, applying them in our research. In the previous work \cite{AhmedFrontiers2021}, a fusion model was developed that combined data-driven models (MSTCN++ \cite{Li2023}, Transformer \cite{VaswaniNeurIPS2017}) and prior knowledge-driven models (HMM \cite{Rabiner1989}, rule-based decision tree) to learn the complex bi-directional inter-state relationship between the segments. This work demonstrated the efficacy of action segmentation for automated movement quality assessment. However, the number of states is greater in the previous home-based setup compared to the ASAR dataset, and the state relationships are bi-directional, making the overall structure more complex.

 \begin{figure}[]
\begin{center}
\centerline{\includegraphics[width=1.1\columnwidth]{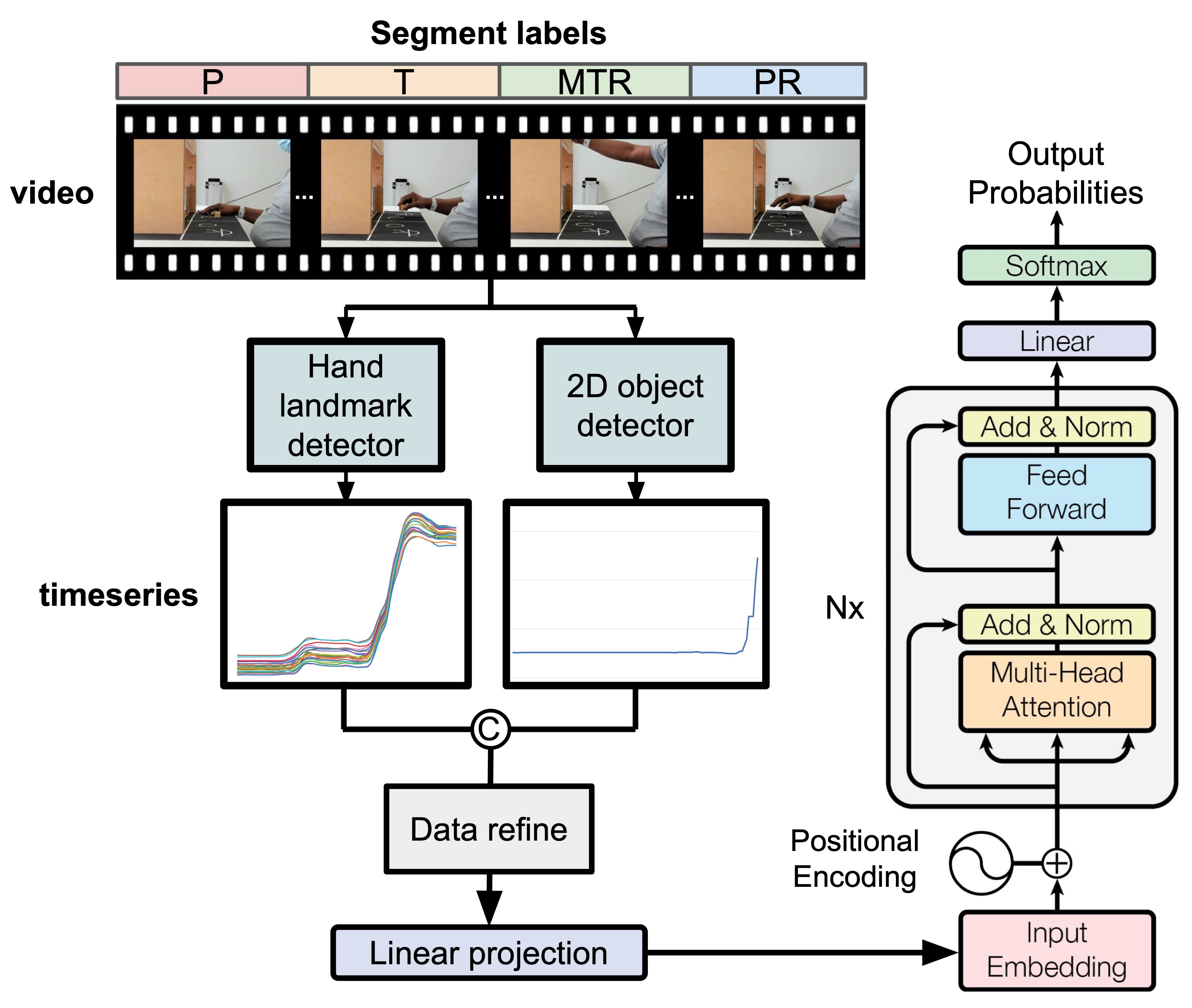}}
\caption{Overview of the three-phase process for temporal keypoint-based video action segmentation using a transformer model. This includes the detection of keypoints (such as object centers and hand landmarks), the refinement of the detected outcomes, and the use of the refined time-series data to train a transformer-based model for accurate timestamp prediction.}
\label{fig:Overview}
\end{center}
\end{figure}

\begin{figure*}[]
  \centering
  \includegraphics[width=2\columnwidth]{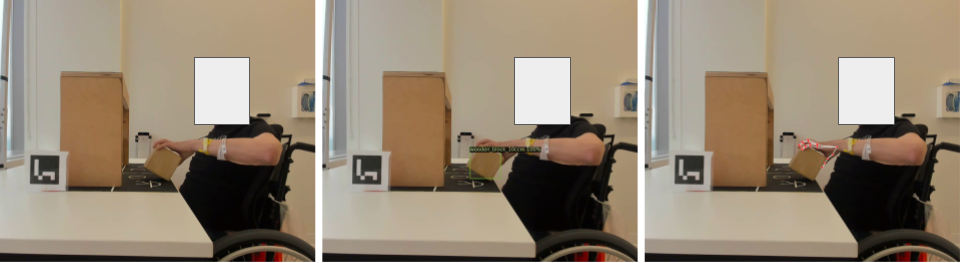}
  \caption{The results of object and hand landmark detection on a sample frame extracted from a video. The left image shows the input frame, the middle image displays the object detection results obtained by TridentNet, and the right image visualizes the 21 hand keypoints extracted using MediaPipe.}
\label{fig:hand_object_detection}
\end{figure*}

\section{Method}

Our research is organized into three primary phases. First, we detect keypoints, including the center coordinates of objects and hand landmarks, from the provided videos. Next, we refine the outcomes to prepare them for input into a temporal segmentation model. Finally, we use the refined time-series data to train the model, enabling it to predict timestamps accurately. Each of these phases is executed sequentially and requires a detailed understanding of the data. The overall process is illustrated in Figure \ref{fig:Overview}. As shown in Figure \ref{fig:Overview}, the process begins with detecting keypoints in the video, followed by refining these points to prepare the data for temporal segmentation. The refined time-series data is then used to train the model, enabling accurate timestamp predictions. This structured approach ensures that each phase builds upon the previous one, ensuring robust segmentation performance.

\subsection{Keypoint detection}

In this study, data collection involved capturing stroke patients' activities from multiple angles, with a total of 12 interacting objects as described in \cite{AhmedICMI2023}. These objects varied in size, ranging from 10 cm wooden blocks to marble balls. These datasets were used to fine-tune TridentNet \cite{LiICCV2019} which is pre-trained on MS-COCO \cite{Lin2014}. We selected this 2D object detection model because its scale-aware training scheme makes it robust in recognizing small objects. Using the fine-tuned model, we inferred the locations of objects and, during the inference process, extracted only the center coordinates from the bounding boxes obtained.


Regarding hands, it is important to obtain the positional information of key joints of the hand individually. Therefore, we utilize Google's MediaPipe \cite{Lugaresi2019} to obtain hand landmark information through only the inference process without additional training. This model identifies the positions of hands participating in activities within image frames and provides 21 finger joint coordinates. 

Figure \ref{fig:hand_object_detection} shows the coordinate position information obtained using these methods. From the target object, a single coordinate is obtained, whereas 21 coordinate values are extracted from the hand. The coordinates obtained in this manner serve as the feature vectors for the temporal segmentation model. Since the results of existing detection models are not entirely accurate, there are cases where detection does not occur for certain frames or where objects are misclassified. This can lead to incomplete sequences when processing with segmentation models or other time-series methods. Therefore, refining the data is necessary for this purpose.

\subsection{Refining detected temporal keypoint data}
Through the object detection model and the hand landmark detection model, we obtained the x and y coordinates of the keypoints. After fine-tuning TridentNet \cite{LiICCV2019} on the given dataset, the object detection accuracy results for the three views were 83.83, 85.36, and 61.79, respectively. For some objects, the model shows an accuracy of 99\%, while for others, it does not even reach 30\%. Therefore, using the object detection results as input data for the time-series segmentation model is not feasible. The experiments related to object detection are further discussed in the experiment section. Similarly, when using the Mediapipe model \cite{Lugaresi2019} for finger joint coordinate detection, there were occasions where the 21 finger joint coordinates were not properly detected due to occlusion.

To address this issue, we decided to utilize our prior knowledge that there is always only one object in each frame for object detection. If the target object was not detected, we used the center coordinates of the object with the highest classification score among the other detected objects as alternative coordinate data. By utilizing this approach, even if the object detection model is not excellent at classification, we can still utilize its localization capabilities. This aligns with our goal of obtaining the target object's trajectory by using the positional information. Next, we examined the object center coordinate data and hand landmark coordinate data, excluding any data with more than 25\% missing values. Even after filtering, some data still had missing values, which we addressed using nearest-neighbor interpolation. We then applied the Savitzky-Golay filter \cite{savitzky1964smoothing} to handle outliers and smooth the data. The processed data were then used as input for the time-series segmentation model. Leveraging the temporal context of the time-series data, this approach allowed us to recover missing data for certain objects, which might have been undetected due to occlusion or other issues.


\subsection{Temporal Keypoints segmentation}

To perform segmentation using temporal keypoint data, we designed the model using the encoder of the vanilla transformer \cite{VaswaniNeurIPS2017}. Specifically, we concatenated the y coordinates of the finger joints and the center y coordinates of the object. Each frame's joint coordinates are first added with positional encodings, and the resulting values are then transformed into Query (Q), Key (K), and Value (V) through linear projection. After that, positional encoding is added. The attention scores were derived by computing the dot product between the query vector and all key vectors. Subsequently, as depicted in Equation 1, the softmax function was applied to the dot product outcomes, facilitating the update of the value vectors.
\begin{equation}
    \text{Attention}(Q, K, V) = \text{softmax}\left(\frac{QK^T}{\sqrt{d_k}}\right)V
\end{equation}
This operation is efficiently performed using the multi-head attention mechanism which measures the similarity between tokens. By finding the relationship among tokens, the model learns which temporal instances belong to the same action label and which do not. The resulting representation is then processed through a linear layer, succeeded by a softmax operation to compute the loss. We used cross-entropy loss as the loss function. When calculating the loss, zero padding added during data preprocessing is excluded from the computation. 

In addition to the model composed solely of transformers, we also used a model combining a transformer as the encoder with an LSTM \cite{Hochreiter1997} to see if it captures task-relevant feature information more effectively. We will later compare this with a model that uses only LSTM for segmentation.

\begin{figure}[tb]
  \centering
  \includegraphics[width=1\columnwidth]{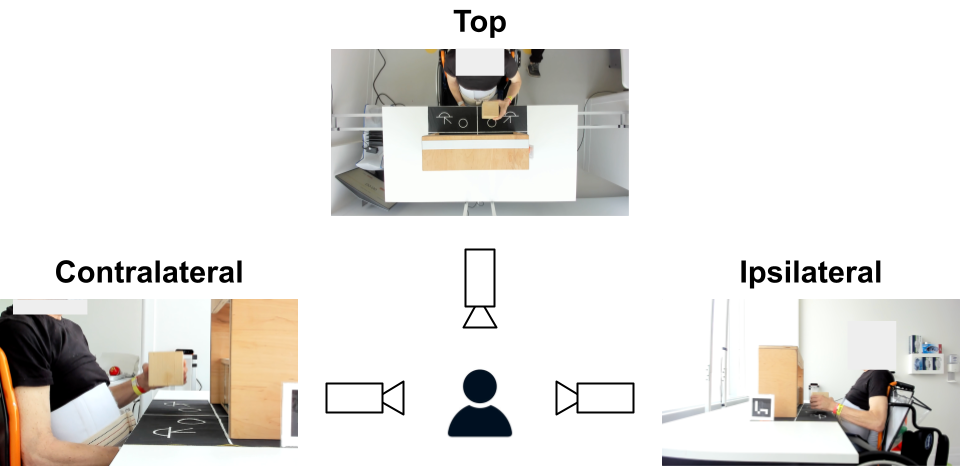}
  \caption{Camera settings used for collecting video data. Three cameras are positioned orthogonal to each other. Data from the left and right cameras is classified based on the patient's hand usage: palm-facing data is contralateral while back-of-the-hand-facing data is ipsilateral.}
\label{fig:camera_views}
\end{figure}

\section{Experiments}

\begin{table*}[]
  \caption{Object detection results for ASAR datasets.
  }
  \label{tab:trident_res}
  \centering
  \begin{tabular}{@{}lccc@{}}
    \toprule
    Objects     & Acc (Top) & Acc (Contralateral) & Acc (Ipsilateral) \\
    \midrule
    wooden block 10cm & 99.34 & 99.70 & 97.51 \\
wooden block 7.5cm & 96.79 & 97.03 & 85.71 \\
wooden block 5cm & 95.54 & 97.69 & 72.08 \\
wooden block 2.5cm & 93.80 & 94.43 & 53.07 \\
cricket ball & 92.65 & 97.39 & 86.50 \\
sharpening stone & 70.21 & 90.16 & 42.22 \\
tumbler & 96.80 & 98.39 & 96.80 \\
thick alloy tube & 92.92 & 98.70 & 94.74 \\
thin alloy tube & 51.47 & 85.03 & 65.89 \\
washer & 81.68 & 34.17 & 4.61 \\
ball bearing & 58.18 & 47.37 & 3.14 \\
marble & 76.60 & 84.24 & 39.21 \\
\midrule
average & 83.83 & 85.36 & 61.79 \\
  \bottomrule
  \end{tabular}
\end{table*}

\begin{table*}[]
  \caption{Frame-wise segmentation accuracies on three view datasets.
  }
  
  \label{tab:segmentation_comparisons}
  \centering
  
  \begin{tabular}{@{}lcccc@{}}
    \toprule
    model       & \#params & Acc (Contralateral) & Acc (Top) & Acc (Ipsilateral) \\
    \midrule
    LSTM1       & 0.28M & 74.83$\pm$ 2.16 & 73.53$\pm$ 1.44 & 80.15$\pm$ 1.39 \\
LSTM3       & 1.3M  & 74.71$\pm$ 1.76 & 74.47$\pm$ 0.66 & 81.02$\pm$ 1.11 \\
Trans3      & 0.25M & 79.78$\pm$ 2.00 & 84.20$\pm$ 0.59 & 82.93$\pm$ 1.62 \\
Trans6      & 0.5M  & 81.81$\pm$ 1.18 & 83.38$\pm$ 2.42 & 82.99$\pm$ 1.86 \\
Trans10     & 0.84M & 83.34$\pm$ 2.86 & \textbf{84.31$\pm$ 0.67} & \textbf{83.08$\pm$ 0.62} \\
Trans3LSTM1 & 0.65M & 79.91$\pm$ 4.86 & 83.08$\pm$ 1.74 & 82.21$\pm$ 1.82 \\
Trans6LSTM1 & 0.9M  & 84.15$\pm$ 0.65 & 84.06$\pm$ 1.43 & 82.99$\pm$ 1.72 \\
Trans3LSTM3 & 1.7M  & \textbf{84.82$\pm$ 2.39} & 82.26$\pm$ 0.63 & 82.79$\pm$ 0.82 \\
  \bottomrule
  \end{tabular}
\end{table*}

\subsection{Datasets}
We used the videos provided in the ASAR Dataset \cite{AhmedICMI2023} as our data. The dataset was generated by recording and annotating the videos of patients as they performed 19 standardized Action Research Arm Test (ARAT) assessment tasks in a clinical setup. Patients were required to move target objects from their original positions to specific locations using various techniques such as Grasp, Grip, and Pinch. This process was captured from three different views.

For object detection, we split the train and test set according to the methods presented in previous research \cite{AhmedFrontiers2021}. We then performed inference on patient groups that were not used for fine-tuning and used this data for the time series segmentation model. Experiments of time-series segmentation were conducted separately for different views: ipsilateral (impaired side of the patient), contralateral (opposite of impaired side), and the top (see Figure \ref{fig:camera_views}). All views include hand keypoint data, and contralateral data has additional information on object position since contralateral data tends to provide clear visibility of the object. Therefore, contralateral data consists of a total of 22 channels, including 21 finger joint coordinates and the object, whereas the top and ipsilateral data are composed of 21 channels corresponding to the number of hand landmarks. We used only the Y-coordinate data from the important parts' positional information obtained in this manner.

The ground truth segment labels consist of four parts for each activity: Initiation and progression (IP), Termination (T), Manipulation and Transportation (MTR), and placement and release (PR). The IP segment starts when the patient's hand begins to move and ends when the hand is positioned in the object-oriented space. T starts when IP ends and finishes when the object is lifted off the table. MTR begins once the object is lifted, and the final sub-activity, PR, starts when the object is near the target space. PR ends when the object is fully released from the hand. In the top view data, all four labels are used. However, for the other two views, only the IP, T, and MTR labels are used for training since the latter part of the videos is excluded from the training process.

The initially collected videos include data from 106 patients performing 19 different tasks. However, due to poor detection performance, during the preprocessing stage, which involved using the results of object detection and hand landmark detection as inputs for the time-series segmentation model, the amount of usable data was reduced. For the three types of views, the train set and test set were split based on patients, with the same 7 patients' data being used as the test set across all three datasets. For the contralateral data, the train set included 50 patients with a corresponding 277 data, while the test set included 42 data. For the top data, the train set consisted of 491 data from 58 patients, and the test set consisted of 70 data from 7 patients. For the ipsilateral data, the train set comprised 564 data from 59 patients, and the test set comprised 79 data from 7 patients. The trainset was further split for k-fold cross-validation \cite{Kohavi1995}, with 20\% of the validation set.

We standardized the total sequence length of data to 300 for model training. Image frames were extracted from the videos at 30 fps. If the total number of frames exceeded 300, downsampling was applied. If it was less than 300, zero padding was added.


\begin{figure*}[]
  \centering
  \includegraphics[width=2.1\columnwidth]{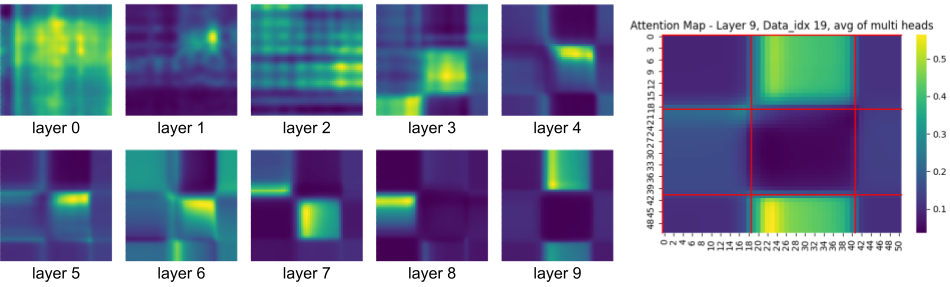}
  \caption{The attention map for contralateral view data. The 10 images on the left show the attention scores of a single head across each of the 10 encoder layers of the Trans10 model. These images reveal how the attention scores evolve with increasing layer depth. The right image presents the average attention scores across the 8 heads of the multi-head attention module in the final encoder layer. This visualization highlights the aggregated attention patterns after processing through all encoder layers.}
\label{fig:attention_maps}
\end{figure*}

\subsection{Experimental Settings for object}

To obtain the center coordinates of the target object, we used a TridentNet pre-trained on the MS-COCO dataset provided by Detectron2. For additional training on the ASAR dataset, we used a learning rate of 0.0025, and trained the model for 60,000 iterations. During training, the learning rate was reduced by a factor of 0.1 at the 47,000th and 55,000th iterations. Table \ref{tab:trident_res} shows the accuracy results of object detection for 12 different objects when using TridentNet. The training was conducted on datasets from three types of views. Overall, the model showed good performance on larger objects such as wooden box 10cm, tumbler, and thick alloy tube. However, its performance was significantly poorer for smaller objects, such as washers, ball bearings, and marbles. Specifically, in the ipsilateral view, accuracy was notably low for these smaller items. The reason for the low detection performance is that the camera is directed toward the back of the hand, so once the object is grasped by the hand, it is completely obscured by the hand.

  
  

\subsection{Experimental Settings for temporal segmentation}
We conducted a series of experiments comparing three types of models: LSTM, Transformer, and combinations of both. Each model was tested with varying numbers of layers. Specifically, for the combined models, we utilized the Transformer model as the initial component, feeding its output features into the subsequent LSTM model. The model names listed in Table \ref{tab:segmentation_comparisons} include the model type and the number of layers used. For instance, "LSTM" denotes a pure LSTM model, while "Trans" refers to a Transformer model. Numbers following each model name indicate the layer count; for example, "Trans3LSTM1" represents a model with a Transformer consisting of 3 encoder layers followed by an LSTM with 1 layer.

In our experimental setup, we employed cross-validation to ensure the robustness of our results. Table \ref{tab:segmentation_comparisons} provides the mean and standard deviation of accuracy metrics computed across five folds. Accuracy was measured frame-wise by determining the proportion of correctly labeled keypoints relative to the total number of keypoints, which corresponds to the number of video frames. All models were trained for 500 epochs with a learning rate of 0.001 using the Adam optimizer \cite{KingBa15}. For the transformer models, the embedding dimension was set to 128, and the multi-head attention mechanism utilized 8 attention heads. The LSTM models had a hidden dimension of 256. For the evaluation metric, we used frame-wise accuracy.

\begin{figure*}[]
  \centering
  \includegraphics[width=2.1\columnwidth]{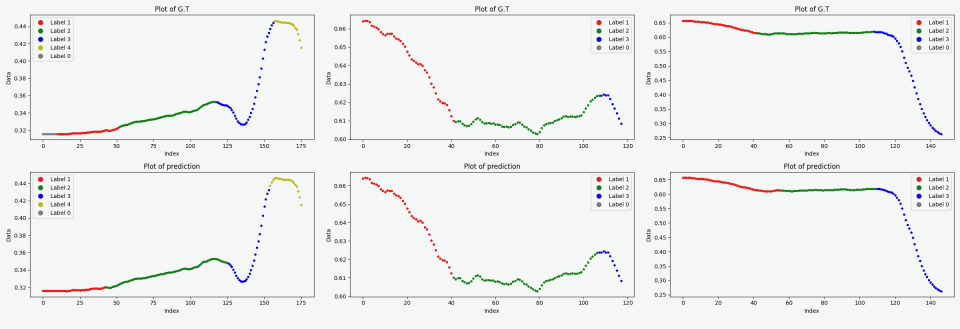}
  \caption{Qualitative results of temporal segmentation. The top row of plots displays the actual segment labels of the data, while the bottom row presents the predicted results. The leftmost plot is a sample from the top view data, the middle plot shows an example from the contralateral view data, and the rightmost plot represents the ipsilateral view data. These plots provide a visual comparison between the ground truth segmentations and the model's predictions across different perspectives.}
\label{fig:plot}
\end{figure*}

\section{Results}

We observed an overall improvement in performance as the number of Transformer encoder layers increased in the model, which can generally be attributed to the self-attention mechanism inherent in the Transformer architecture. Figure \ref{fig:attention_maps} visualizes the attention score from one head of the multi-head attention mechanism for a single sample. Up to layer 2, the attention map does not provide meaningful information for the segmentation task. However, from layer 3 onward, the attention map reveals increased similarity within specific regions, forming distinct blocks. The red line in the right image of Figure \ref{fig:attention_maps} represents the ground truth segment label. Comparing the actual labels with block boundaries indicates the crucial role of the transformer's self-attention mechanism in segmenting the one-dimensional time-series data. This visualization provides additional insight into how the number of layers contributes to performance improvement, complementing the accuracy metrics from the experimental results.

The performance improvement with increasing Transformer layers was particularly pronounced for the contralateral view, which had the smallest amount of data. This indicates that deeper models can be more beneficial for final predictions when working with limited data. Furthermore, combining a few Transformer encoder layers with LSTM modules improved performance over using either the Transformer or LSTM alone, despite Transformers having fewer trainable parameters. This indicates that the addition of Transformer layers can offer a significant performance boost even with a small dataset.

The effectiveness of temporal segmentation is further demonstrated in Figure \ref{fig:plot}, which presents plots of a single activity performed by one patient from three different views. The x-axis represents time, while the y-axis denotes the y-coordinate of the wrist in each frame. Wrist position coordinates, plotted in different colors, highlight distinctions between various actions.

By analyzing the trajectories of keypoints from multiple videos of a single scene, we were able to identify which view most effectively identifies the boundaries between segment labels. This analysis of temporal keypoint data allowed us to assess the strengths and weaknesses of each view dataset, offering insights into how to optimally utilize this information.

\section{Conclusion}

Through this research, we have demonstrated an effective framework for structuring models to handle sophisticated movements common in physical therapy by breaking them down into smaller, manageable tasks based on domain-specific prior knowledge. This approach has proven particularly beneficial when working with limited real-world data, showcasing the feasibility of achieving performance levels suitable for clinical applications. Our findings emphasize the value of integrating an understanding of movement biomechanics into the model design, enabling precise temporal segmentation even with small datasets. Additionally, the use of refined keypoint data has demonstrated how focusing on patient-specific movements enhances the efficiency of segmenting and analyzing relevant actions.

We also conducted a comprehensive analysis of the influence of encoder layer depth in Transformers on time-series segmentation accuracy, underscoring the critical role of model architecture in achieving reliable results. Furthermore, our approach includes a refinement process following the detection of object and hand locations, which has addressed some limitations in detection accuracy and improved the quality of the data used for temporal segmentation.

Looking ahead, this framework has broader applicability in monitoring specific actions within small datasets, such as those capturing infants, pets, or elderly individuals via home cameras. The potential to extend this technology to other domains highlights its versatility and impact beyond stroke rehabilitation.

For future research, there is significant potential to explore advanced techniques for enhancing segmentation performance by incorporating data from multiple camera angles or additional sensor modalities. Utilizing these diverse data sources could provide a richer context and further improve the accuracy and robustness of segmentation, offering more precise insights into complex movement patterns.


\section*{Acknowledgements}

This work was supported by NSF grant 2230762.

{\small
\bibliographystyle{ieee_fullname}
\bibliography{egbib}
}

\end{document}